\documentclass[runningheads]{llncs}
\usepackage[T1]{fontenc}
\usepackage{graphicx}
\usepackage{booktabs} 
\usepackage{algorithmic}
\usepackage{algorithm}
\usepackage{amsmath}
\usepackage{adjustbox}
\usepackage{color}
\begin{document}
\title{Generalization Gap in Data Augmentation: Insights from Illumination}
%
%
\author{Jianqiang Xiao\inst{1,2} \and
Weiwen Guo\inst{3}\and
Junfeng Liu\inst{1}\and
Mengze Li\inst{4}}
\authorrunning{F. Author et al.}
%
\institute{South China University of Technology, Guangzhou, China 
\and Hitachi Elevator (China) Co., Ltd. Guangzhou, China
\and Hitachi Building Technology (Guangzhou) Co., Ltd. Guangzhou, China 
\and Research Center for Frontier Fundamental Studies, Zhejiang Lab, Hangzhou, China}
\maketitle              
\begin{abstract}
In the field of computer vision, data augmentation is widely used to enrich the feature complexity of training datasets with deep learning techniques. However, regarding the generalization capabilities of models, the difference in artificial features generated by data augmentation and natural visual features has not been fully revealed. This study introduces the concept of "visual representation variables" to define the possible visual variations in a task as a joint distribution of these variables. We focus on the visual representation variable "illumination", by simulating its distribution degradation and examining how data augmentation techniques enhance model performance on a classification task. Our goal is to investigate the differences in generalization between models trained with augmented data and those trained under real-world illumination conditions. Results indicate that after applying various data augmentation methods, model performance has significantly improved. Yet, a noticeable generalization gap still exists after utilizing various data augmentation methods, emphasizing the critical role of feature diversity in the training set for enhancing model generalization.

\end{abstract}
\section{Introduction}
Over the past ten years, there has been a significant revolution in computer vision field. The advancement mainly belongs to deep learning techniques, particularly the utilization of Convolutional Neural Networks (CNNs)~\cite{ref1,ref2,ref3,ref4} and Transformer~\cite{ref5,ref6} architectures. By emulating creatural visual systems, CNNs stack convolutional layers and pooling layers, while Transformers utilize self-attention mechanisms~\cite{ref7} to handle long-range image dependencies effectively. With all these efforts, researchers made significant progress in image classification~\cite{ref1,ref2,ref3,ref4}, object detection~\cite{ref8,ref9}, and semantic segmentation~\cite{ref10,ref11}. Nowadays, computer vision expands its applications in industry~\cite{ref12}, healthcare~\cite{ref13}, and transportation~\cite{ref14}, improving convenience and quality of people’s lives.

In the realm of deep learning-based computer vision algorithms, the quality and variety of data significantly impact the generalization of visual models~\cite{ref15,ref16}. Despite the importance of real-world data collection, challenges such as scene diversity, labeling expenses, and privacy issues hinder the ability of extensive datasets to fully capture all visual characteristics. Through techniques such as geometric modifications, color channel adjustments, and filter incorporation~\cite{ref17,ref18}, data augmentation effectively enhances visual feature diversity in datasets, improving training model generalization capabilities. Furthermore, data augmentation serves as a crucial method to prevent overfitting, especially when dealing with limited data. Consequently, data augmentation has become a fundamental component in the training process of computer vision application projects~\cite{ref19,ref20,ref21}.

The visual diversity in datasets can be broadly divided into two categories: changes in the intrinsic properties of the recognition objects (e.g., the variety of vehicles in autonomous driving scenarios~\cite{ref22}), and changes indirectly caused by external environmental factors (e.g., different weather conditions~\cite{ref23} in self-driving). The visual diversity in the dataset ensures the model preserves robust recognition capabilities across various scenarios. However, when the distribution of visual characteristics is uneven, potentially diminishes the model's performance across various environmental conditions~\cite{ref24}. When data distribution imbalances occur, data augmentation is widely used as an effective method~\cite{ref25,ref26,ref27}. However, a discernible gap remains between augmented images and those captured in the physical world, highlighting under certain extreme conditions, such as adverse weather, the artificial features produced by data augmentation could potentially impair model performance~\cite{ref28,ref29}. This has prompted a reevaluation of data augmentation, whether synthetic, non-realistic pixel-wise feature characteristics might undermine a model's generalization in real-world scenario. Our study involves controlling illumination settings in a classification task to compare model performance under real-world and data augmentation datasets, aiming reveal the effectiveness and limitations of data augmentation. 

\begin{figure}[!t]
	\centering
	\includegraphics[width=4.5in]{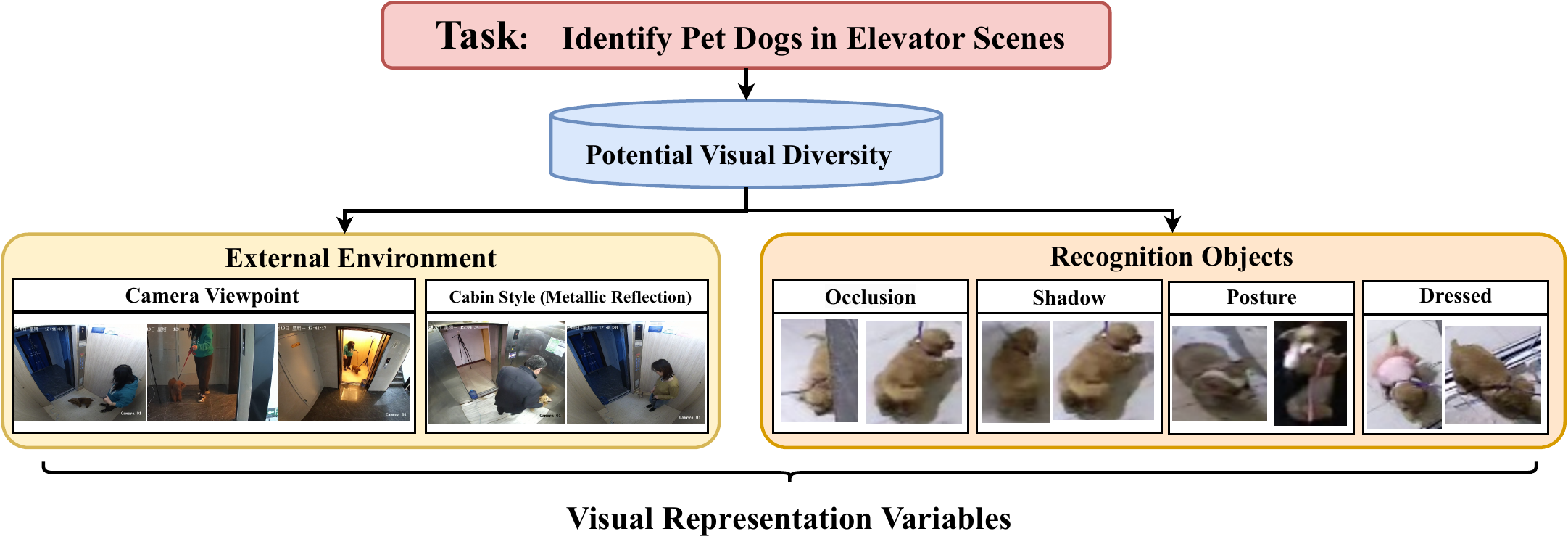}
	\caption{Visual representation variables decomposition guided by task prior knowledge.}
	\label{fig1}
\end{figure}

In our study, we introduce the concept of "visual representation variables" to define the potential visual changes in a task as a joint distribution of these variables (Fig.~\ref{fig1}). We focus on isolating a single visual representation variable, "illumination," to comprehensively study its effects. "Illumination" was chosen because it can be quantitatively measured and is relatively easy to replicate in data augmentation. By controlling illumination settings in a classification task, we compare model performance under real-world and data augmentation conditions to reveal the effectiveness and limitations of data augmentation.

\textbf{Our main contributions are as follows:}

\begin{itemize}
	\item{We validate that the model's generalization ability suffers a devastating impact when the illumination environment degrades into a singular distribution;}
	\item{By using a gray card to measure scene illumination mapping and optimizing color data augmentation parameters through Bayesian optimization, we achieve significant improvements in the generalization ability of datasets with singular illumination distribution;}
	\item{We demonstrate a significant generalization gap between models trained with data augmentation and those trained with real-world data. This emphasizes the limitations of data augmentation in replicating real-world visual features and underscores the necessity of carefully designed datasets.}
\end{itemize}

\section{Related works}
\subsection{Domain Generalization}
Domain generalization aims to enhance a model's performance on unseen data (target domain), particularly when there is a distributional discrepancy between the training data (source domain) and test data. The main strategies include domain alignment~\cite{ref30} to minimize distributional differences between source and target domains, meta-learning~\cite{ref31} which leverages learning across tasks to boost generalization capabilities, data augmentation~\cite{ref32,ref33} that introduces sample visual diversity to enhance model robustness and ensemble learning~\cite{ref34} which integrates multiple models to optimize overall performance. The issue of domain shift in domain generalization can be decomposed into differences in the distribution of a series of "visual representation variables" between the source domains and target domains. Our study focuses on data-level augmentation as it directly enriches the visual features of training data without the need to alter network structures or training strategies, closely aligning with our goal of exploring the impact of illumination environments on model generalization.

\subsection{Data Augmentation}
Data augmentation is a crucial technique for enhancing the generalization ability of deep learning models. Common methods, such as geometric transformations~\cite{ref35}, color space adjustments~\cite{ref36}, and random cropping~\cite{ref37}, expand the representation of visual features and strengthen the model's generalization ability to recognize unseen data. Specifically, adjustments in the color space simulate different lighting conditions, directly influencing the model's adaptability to changes in illumination~\cite{ref38}. Although current research on data augmentation mainly focuses on its efficacy in enhancing model generalization, there has been limited exploration of the holistic impact of data augmentation techniques in simulating complex environmental changes, such as illumination settings~\cite{ref39}. Our study focuses on illumination settings, investigating the generalization effects of models by constructing distributions of real lighting and corresponding data-augmented distributions. This approach aims to reveal the potential disparities in generalization capabilities between augmented datasets and real-world datasets.

\section{Experimental Framework and Data Preparation}
\subsection{Recognition Targets}
\begin{figure}[!t]
	\centering
	\includegraphics[width=4.5in]{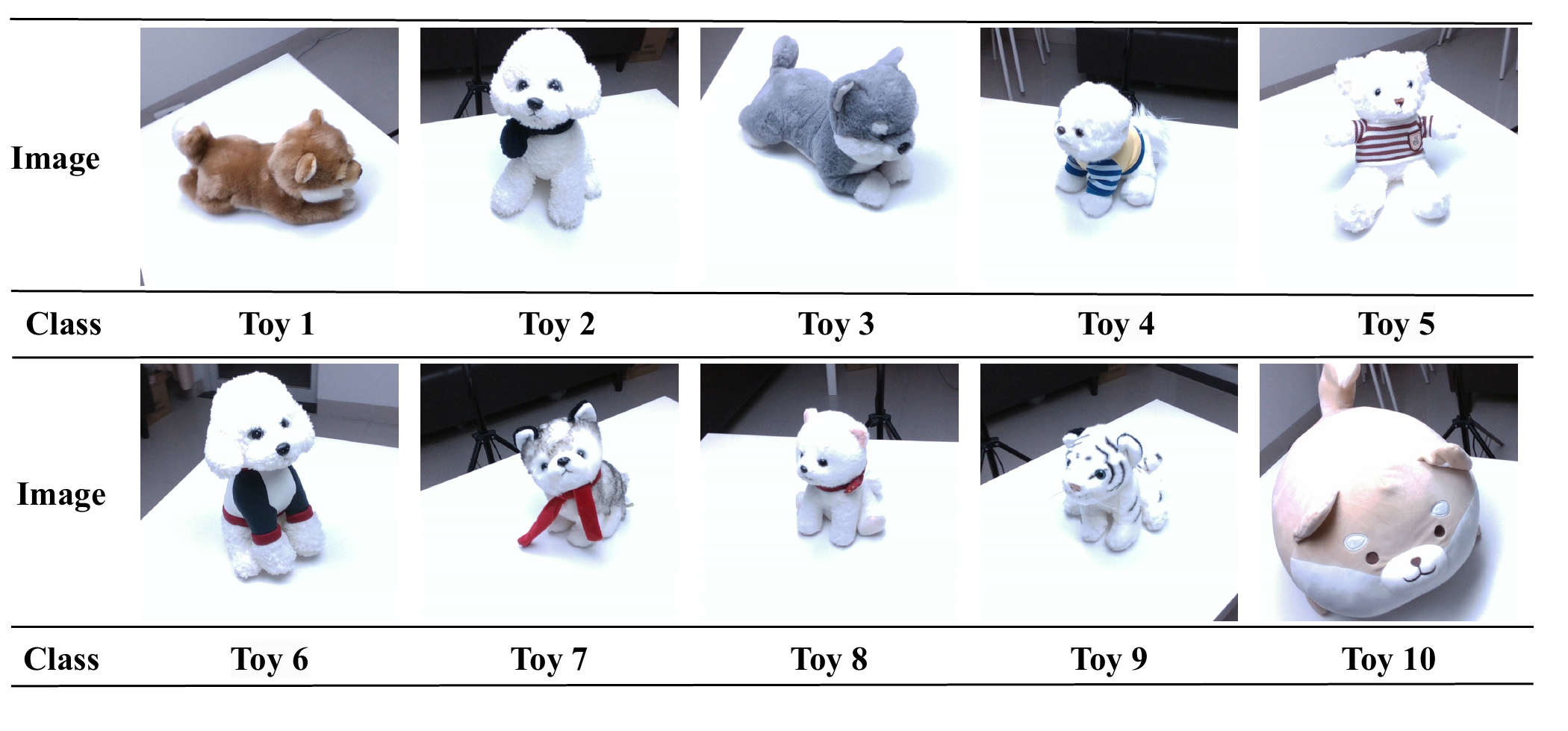}
	\caption{An assortment of 10 distinct toy dogs serves as recognition targets in our classification task. The variety in their visual features, such as shape, color, fur texture, and attire, highlights the complexity of our dataset and assesses the classification models' ability to distinguish visual differences from subtle to pronounced.}
	\label{fig2}
\end{figure}

To verify the universality of our insights, we selected basic image classification as our target task. As shown in Fig.~\ref{fig2}, the collection of objects comprises ten different toys designed to challenge vision models' ability to recognize a wide range of visual aspects. The diversity in shape, color, and texture of these toys spans from easily recognizable to complex features, thoroughly testing the models' capabilities in processing subtle visual differences. This systematic selection strategy evaluates the models' recognition ability to handle visual complexity and reveals the delicate differences in visual features during the generalization process, enhancing our understanding of the image classification mechanism.

Given that this study primarily investigates the impact of data augmentation techniques on enhancing model generalization capabilities, focusing especially on the challenge of the "illumination" variable compared to real datasets, we aim to dive into how models adapt to visual stimuli across different data augmentation methodologies. By selecting toys with distinct visual characteristics as recognition objects, we established a foundation for subsequent research into data augmentation and model generalization.

\subsection{Illumination Environment Setting}
\begin{figure}[!t]
	\centering
	\includegraphics[width=4.5in]{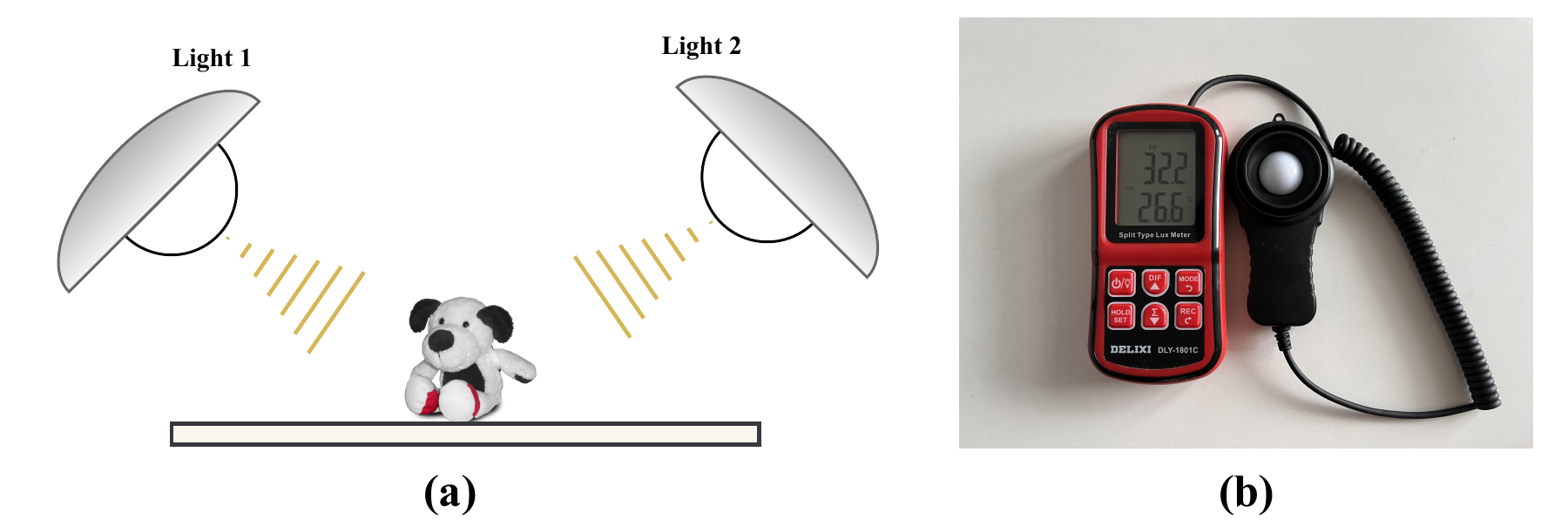}
	\caption{(a) A dual light source setup with supplementary lamps placed at 45-degree angles to ensure balanced illumination. (b) The light intensity meter for precise measurement of illumination conditions.}
	\label{fig3}
\end{figure}

To examine the impact of illumination as a critical visual characteristic variable on model generalization, we designed our experimental environment as illustrated in Fig.~\ref{fig3} (a). Our goal was to create a stable and uniform lighting environment, crucial for ensuring the integrity and reliability of our results. For this purpose, two fill lights were positioned at 45 degrees on either side of the experimental platform. This arrangement created a balanced and even dual light source environment, effectively eliminating potential shadows or irregular illumination during the data-taking process. These fill lights were adjustable, capable of emitting light at three different color temperatures and allowing for controlling the intensity via remote control, thus establishing a uniform dual-source lighting condition.

As an integral part of this setup, we utilize a lux meter (as shown in Fig.~\ref{fig3} (b)) to provide precise quantification of the illumination environment. This instrument is crucial for precision measurements of light intensity in various illumination settings, enabling us to describe the attributes of each scene quantitatively. Carefully designed illumination environments and quantitative measurements form the foundation for constructing illumination distributions in our research, ensuring that variations in lighting can be precisely controlled and accurately quantified. All these illumination environment settings built a comprehensive experimental environment with dual controllable light sources, providing a trustworthy platform for our data collection process.

\subsection{Data Preparation}
\subsubsection{Training Set}
In order to thoroughly investigate the impact of the visual variable, "illumination", on the generalization capabilities of visual models, we purposefully constructed two training sets. The Full Spectrum Illumination Dataset (FSID) encompasses a range of illumination distributions, incorporating variations in light color and intensity. This representation ensures a uniform distribution of the "illumination" variable in our study. In contrast, the Singular Illumination Dataset (SID) simulates a narrowed range of illumination by selecting specific median illumination settings from FSID, transitioning from a broad, uniform distribution to a degraded, singular distribution.

\subsubsection{Full Spectrum Illumination Dataset (FSID):}

\begin{table*}[ht]
	\centering
	\caption{Illumination settings with different levels of intensity and color temperature, where illumination intensity was measured using the light intensity meter in Fig. ~\ref{fig3} (b), ensuring an error margin within ±20 lux. And the color temperature was indirectly measured using a gray card under different illumination settings. An average value was calculated from 100 images.}
	\label{tab1}
	\begin{adjustbox}{width=1\textwidth,center}
		\begin{tabular}{@{}ccccccc@{}}
			\toprule
			\textbf{Intensity} & \textbf{-2 level} & \textbf{-1 level} & \textbf{0 level} & \textbf{+1 level} & \textbf{+2 level} \\
			\midrule
			\textbf{Warm Light} & 180Lux, 3222K & 540Lux, 3812K & 900Lux, 4205K & 1260Lux, 4388K & 1620Lux, 4205K \\
			\textbf{White Light} & 200Lux, 20397K & 600Lux, 15186K & 1000Lux, 12769K & 1400Lux, 12527K & 1800Lux, 11931K \\
			\textbf{Mixed Light} & 400Lux, 8058K & 1200Lux, 7628K & 2000Lux, 7192K & 2700Lux, 6607K & 3500Lux, 6499K \\
			\bottomrule
		\end{tabular}
	\end{adjustbox}
\end{table*}

\begin{figure}[!t]
	\centering
	\includegraphics[width=4in]{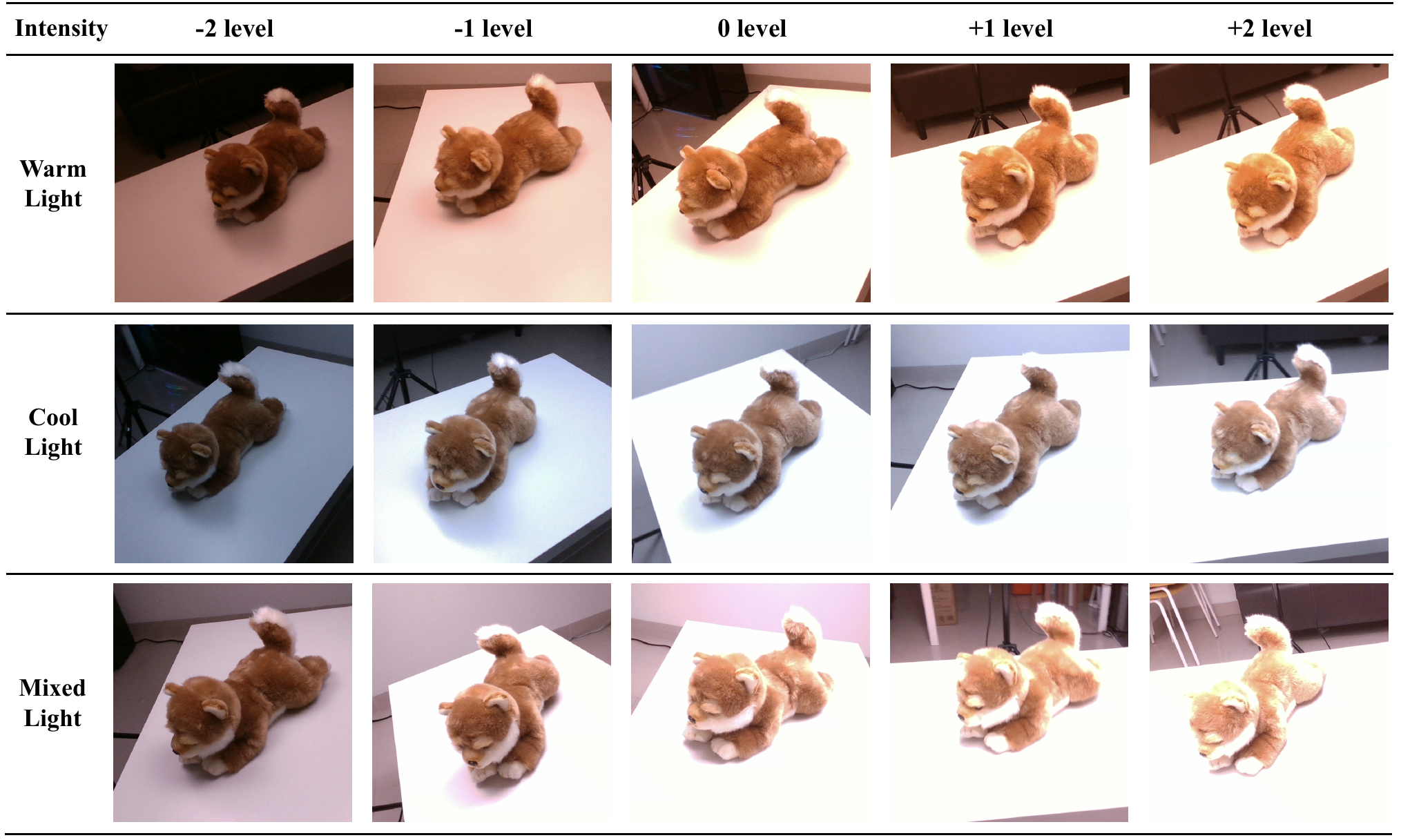}
	\caption{Toy 1 depicted under 15 illumination settings within the FSID, with light colors [Warm, Cool, Mixed] and intensities [-2, -1, 0, +1, +2]. The illumination intensity and color temperature are described in Tab. ~\ref{tab1}.}
	\label{fig4}
\end{figure}

For this dataset, we followed a detailed data collection process under various illumination settings. The illumination attributes include light color and illumination intensity. The light color included [Warm, Cool, Mixed] three different attributes, while the illumination intensity was divided into five distinct levels [-2, -1, 0, +1, 1] by measuring the intensity's scope of different light colors. By forming 15 different illumination settings (3 light colors * 5 illumination intensities), we created a series of diverse illumination variations and performed data-taking based on these settings. For each toy, high-resolution images were captured by high resolution camera under these varied illumination settings, ensuring at least 100 clear, unobstructed images from different angles. Eventually, a set of  15,000 images was collected (3 color temperatures * 5 levels of light intensity * 10 categories * 100 images) as our FSID. Fig.~\ref{fig4} shows images of Toy 1 under 15 different lighting scenarios as an example.

\subsubsection{Singular Illumination Dataset (SID):}
In this dataset, we focused on constructing image data under a singular illumination setting. We chose the middle illumination setting [Cool light, 0 level] listed in Fig.~\ref{fig4} as the target condition, with at least 1500 images of various poses collected around each recognition object to ensure data volume consistency with FSID. The construction of this dataset served two purposes: on one hand, to explore the impact on visual model generalization when visual representation variables are simplified to a singular distribution; on the other hand, to provide a basis for data augmentation study on enhancing model generalization abilities. \\

During the data-taking progress of our study, we used the Intel RealSense D435i camera to capture images of 10 toys from a similar height. All RGB images in our two datasets are clear and unobstructed, with a resolution of 640*640. This consistent data-taking method assures that the primary difference between the FSID and SID training sets is attributed to changes in illumination settings, without introducing additional covariates. This provides a solid basis for further analyzing model performance under different illumination settings as a training set, ensuring the reliability of the research findings.

\subsubsection{Test set:}\label{testset}
Throughout the process of evaluating models' adaptability to changes in illumination, a standardized test set was specifically constructed, characterized by a complexity of lighting variations far exceeding all scenarios within the training sets. The construction of this test set took into account the diverse changes in light color properties and ensured random fluctuations in light intensity across the entire spectrum. Contrasting with the fixed light intensity levels of training set A (five predefined levels for each light color), the test set was designed to cover the complete range of illumination spectrum. Through meticulous adjustments of auxiliary lighting in the testing equipment, a brightness cycle from bright to dark and back to bright was achieved, ensuring a uniform distribution of brightness throughout the range. 

To assess the relationship between model generalization capabilities and the illumination distribution in the training set, we constructed a standardized test set with a range of illumination variations that exceed all scenarios in the training set. The construction of this test set considered the diverse changes in light color properties and ensured random fluctuations in light intensity across the entire scope. Contrasting with the fixed light intensity levels of FSID (five predefined levels for each light color), the test set was designed to cover the complete range of illumination intensity scope. By continuously adjusting the light intensity during the data collection process, and cycling from bright to dark and back to bright, we ensured a uniform distribution of light intensity across the entire range in the test set.

\subsection{Classification Model}
To ensure the experimental findings are widely applicable, we selected a variety of widely-used deep learning models for comparative experiments, including AlexNet~\cite{ref1}, VGG~\cite{ref2}, ResNet~\cite{ref3}, EfficientNet~\cite{ref4}, ViT~\cite{ref5}, and Swin Transformer~\cite{ref6}, including classic CNNs and latest Transformer models. These deep learning models were originally designed to process large, semantically abundant datasets like ImageNet~\cite{ref40}, which provides a comprehensive feature collection in visual tasks.  However, when applying these deep learning models to the relatively simple training sets in our study, all models find it hard to reach convergence during the training process. This is due to the dominant feature complexity gap between our training set and comprehensive datasets such as ImageNet. Finally, we solved this problem by scaling the models to match the capacity of our training data. For CNN models, including AlexNet, VGG, ResNet, and EfficientNet, we adjusted by reducing the number of channels to one-quarter of their original count. For Transformer architectures, with means ViT and Swin Transformer, we made more meticulous modifications. Specifically, we simplified the number of attention heads per layer and reduced the number of hidden units in each layer, aiming to decrease model complexity and computational burden. All these adjustments simplified the model structures while retaining their powerful image processing capabilities. By optimizing the architectures in this manner, we ensured faster model convergence on smaller-scale datasets, preserving the key architecture of their original designs. After confirming that all deep learning models could be trained on FSID and SID with steady loss reduction and ultimately reached convergence, we completed all preparatory work for the following comparative experiments.

\section{Comparative Experiments}
\subsection{Experiment 1: Uniform Distribution vs. Singular Distribution of Illumination Attributes in Training Set}
In our first experiment, we focused on how the distribution of the "illumination" attribute within training datasets affects the generalization performance of deep learning models. Specifically, we utilized the six deep learning models mentioned in Section 3.4 and conducted comparative experiments using the Full Spectrum Illumination Dataset (FSID) and the Singular Illumination Dataset (SID) as the training datasets. We aimed to examine the impact of differences in the distribution of the "illumination" variable on model generalization while keeping all other variables unchanged.

To ensure fairness and consistency for all deep learning models, we applied nearly the same training hyperparameters for all experiments. This included setting the data validation split to 0.2, the batch size to 64, choosing Adam as the optimizer, and setting the initial learning rate to 0.001. During the preprocess stage, we follow the procedural of AlexNet [1] by scaling all training images to 224*224 and making color normalization. All models were trained extensively on FSID and SID until their training loss reached convergence. Notably, CNN-based models typically converged within 10 epochs, while Transformer-based models required more training epochs. Specifically, our ViT model needed 20 epochs, and our Swin Transformer model needed 60 epochs to reach the convergence~\cite{ref41}. After completing all training processes, we evaluated the generalization performance of all models trained on FSID and SID using the test set described in Section ~\ref{testset}. The evaluation metrics included model accuracy, precision, and recall on our test set, the experimental results details were shown in Tab. ~\ref{tab2}.

\begin{table*}[ht]
	\centering
	\caption{Performance comparisons of different models trained on FSID and SID datasets under identical configurations reveal that accuracy declined by 0.67 across the test sets.}
	\label{tab2}
	\begin{adjustbox}{width=1\textwidth,center}
		\begin{tabular}{@{}ccccccccccccccccccc@{}}
			\toprule
			& \multicolumn{3}{c}{\textbf{AlexNet}} & \multicolumn{3}{c}{\textbf{VGG}} & \multicolumn{3}{c}{\textbf{ResNet}} & \multicolumn{3}{c}{\textbf{EfficientNet}} & \multicolumn{3}{c}{\textbf{ViT}} & \multicolumn{3}{c}{\textbf{Swin\_T}} \\
			\cmidrule(lr){2-4} \cmidrule(lr){5-7} \cmidrule(lr){8-10} \cmidrule(lr){11-13} \cmidrule(lr){14-16} \cmidrule(lr){17-19}
			\textbf{Metrics} & \textbf{Acc.} & \textbf{Pre.} & \textbf{Rec.} & \textbf{Acc.} & \textbf{Pre.} & \textbf{Rec.} & \textbf{Acc.} & \textbf{Pre.} & \textbf{Rec.} & \textbf{Acc.} & \textbf{Pre.} & \textbf{Rec.} & \textbf{Acc.} & \textbf{Pre.} & \textbf{Rec.} & \textbf{Acc.} & \textbf{Pre.} & \textbf{Rec.} \\
			\midrule
			\textbf{FSID} & 0.981 & 0.982 & 0.981 & 0.995 & 0.995 & 0.995 & 0.997 & 0.997 & 0.997 & 0.996 & 0.996 & 0.996 & 0.981 & 0.981 & 0.981 & 0.982 & 0.982 & 0.982 \\
			\textbf{SID} & 0.347 & 0.580 & 0.347 & 0.324 & 0.570 & 0.324 & 0.382 & 0.313 & 0.382 & 0.373 & 0.678 & 0.373 & 0.290 & 0.604 & 0.290 & 0.264 & 0.502 & 0.264 \\
			\bottomrule
		\end{tabular}
	\end{adjustbox}
\end{table*}

By analyzing the results showcased in Tab. ~\ref{tab2}, when the illumination distribution degenerates to a singular one, we could clearly recognize the difference of their performance on the test set. Although all models preserved the same training configuration in FSID and SID, the four CNN and two Transformer models exhibited significant performance discrepancies due to their training sets containing different illumination distributions.  The SID only contained a singular illumination distribution [Cool light, 0 level] while the FSID provided a full spectrum distribution with 15 illumination settings. Particularly, all models trained on FSID performed great performance on the test set, however, models received about 0.67 drops in accuracy when trained on SID, with similar declines observed in precision and recall. These results compellingly demonstrate that when the visual variable distribution in training sets degenerated (specifically illumination in our study), this led to a catastrophic decrease in their generalization capabilities.

\subsection{Experiment 2: Statistical Illumination Vector Mapping Augmentation in the Singular Illumination Dataset}

In experiment 1, we confirmed that a singular distribution of illumination significantly reduces the performance of deep learning models. Further observation of the data showed that the color and intensity of lighting have a significant impact on the visual characteristics, especially on the color appearance of training images. Given this observation, we hypothesized that quantifying the pixel-level mapping correlations between different illumination settings could mitigate the decline in model generalization caused by a singular illumination distribution in SID. In experiment 2, we aimed to explore this hypothesis and attempted to address this issue through color channel enhancement methods~\cite{ref42}.

\begin{figure}[!t]
	\centering
	\includegraphics[width=4in]{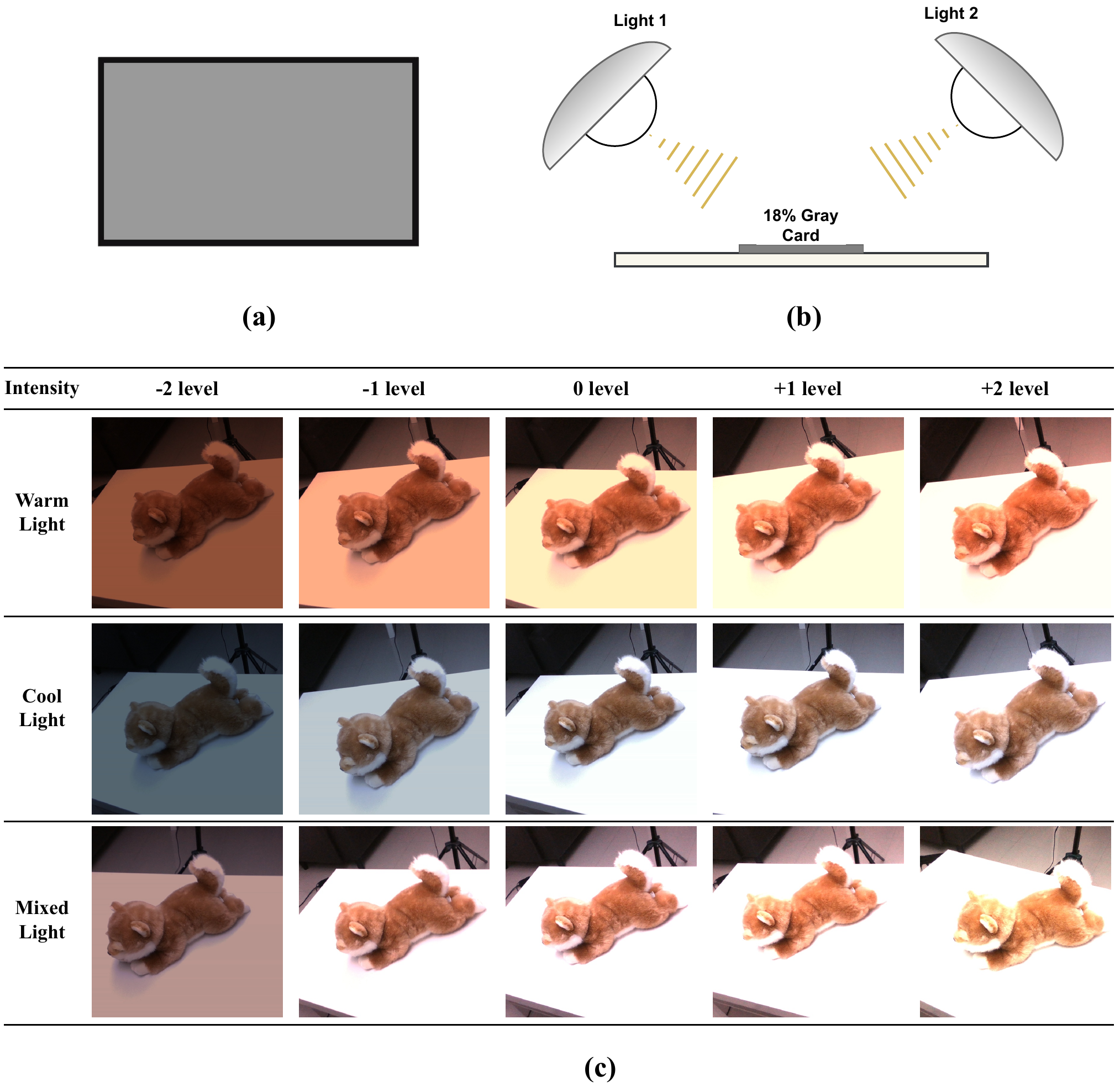}
	\caption{Establishing the illumination settings of FSID to generate extensive illumination vectors for augmenting the SID dataset. (a) 18\% gray card, (b) scene assembled for data collection, and (c) images from the SID dataset of Toy 1, enhanced with illumination vectors under diverse illumination settings (detailed in Tab.~\ref{tab1}). }
	\label{fig5}
\end{figure}

In this experiment, we used an 18\% gray card shown in Fig. ~\ref{fig5}(a) as the subject and replicated the same 15 diverse illumination settings found in the Full Spectrum Illumination Dataset (FSID), as depicted in Fig. ~\ref{fig5}(b). Under these conditions, we captured multiple photographs. From these, we meticulously selected 100 images with consistent imaging quality for further analysis. For each selected image, we calculated the average values of the \(R\), \(G\), and \(B\) color channels under the current lighting conditions, defining it as the environmental illumination vector \(V_{\text{ill}}[R,G,B]\), where the light color \(C\) includes three types (Warm, Cool, Mixed) and light intensity \(I\) is divided into five levels (-2, -1, 0, +1, +2). Then, we calculated the standard illumination environment vector \(V_{\text{ill,SID}}[R,G,B]\) in SID for the [Cool, 0 level] scenario and compared it with the vectors \(V_{\text{ill,FSID}}^{(k)}[R,G,B]\) under the other 14 different lighting conditions. Based on these ratios, we enhanced 15,000 images in SID, selecting 100 images randomly for each lighting condition for processing. The enhancement process was accomplished by applying the corresponding illumination mapping coefficients to the global pixels of the images, thus creating the Illumination Vector Augmentation Dataset (IVAD).

In this experiment, we used an 18\% gray card (shown in Fig. 4(a)) to fill the entire field of view of the camera and replicated the same 15 diverse illumination settings in FSID, as depicted in Fig. 4(b). In this experimental setup, we captured a series of images to calculate the illumination setting vectors. For each illumination setting, we randomly selected 100 images with consistent imaging quality for further analysis. For each selected image, we calculated the average of the \(R\), \(G\), and \(B\) color channels under the current illumination settings, defining it as the environmental illumination setting vector \(V_{\text{illumination}}[R,G,B]\), where the light color \(C\) includes three types [Warm, Cool, Mixed] and light intensity \(I\) is divided into five levels [-2, -1, 0, +1, +2]. Then, we calculated the singular illumination setting vector \(V_{\text{illumination,SID}}^{(1)}[R,G,B]\) in SID for a [Cool, 0 level] scenario and compared it with the vectors \(V_{\text{illumination,FSID}}^{(1 \sim 15)}[R,G,B]\) under the other 14 different illumination settings. Based on the ratios between \(V_{\text{illumination,SID}}^{(1)}[R,G,B]\) and \( V_{\text{illumination,FSID}}^{(1 \sim 15)}[R,G,B]\), we enhanced 15,000 images in SID by multiplying all pixels with the ratio of corresponding illumination setting vectors. For each illumination setting, we randomly selected 100 images, ensuring that the enhanced dataset could match FSID. Through this statistical-based strategy of color channel enhancement method, we created the Illumination Vector Augmentation Dataset (IVAD) based on images in SID.

Fig. ~\ref{fig4}(c) showed the results of Toy1 in IVAD enhanced by illumination mapping enhancement, where the dataset shifted from a singular illumination distribution of [Cool light, 0 level] in SID to 15 types of full-spectrum illumination distributions. This enhancement through color channel-based illumination mapping allowed images in SID, previously constrained by a singular illumination distribution and lacking visual diversity, to present much more complicated visual characteristics. These characteristics were similar to those observed under various illumination settings in FSID. Subsequently, following the setup from Experiment 1, we trained all deep learning models in IVAD. As shown in Table 2, our experimental results indicate that models trained on IVAD achieved an increase of approximately 0.57 in accuracy on the test set compared to those trained on SID. This means that using illumination vector mapping as our data augmentation method could significantly improve models' performance comparing those trained on a singular illumination distribution in SID.

\begin{table*}[ht]
	\centering
	\caption{Comparative analysis of model performance on the SID and IVAD, highlighting the effectiveness of illumination vector-based data augmentation method for improving model generalization capabilities.}
	\label{tab3}
	\begin{adjustbox}{width=1\textwidth,center}
		\begin{tabular}{@{}ccccccccccccccccccc@{}}
			\toprule
			& \multicolumn{3}{c}{\textbf{AlexNet}} & \multicolumn{3}{c}{\textbf{VGG}} & \multicolumn{3}{c}{\textbf{ResNet}} & \multicolumn{3}{c}{\textbf{EfficientNet}} & \multicolumn{3}{c}{\textbf{ViT}} & \multicolumn{3}{c}{\textbf{Swin\_T}} \\
			\cmidrule(lr){2-4} \cmidrule(lr){5-7} \cmidrule(lr){8-10} \cmidrule(lr){11-13} \cmidrule(lr){14-16} \cmidrule(lr){17-19}
			\textbf{Data} & \textbf{Acc.} & \textbf{Pre.} & \textbf{Rec.} & \textbf{Acc.} & \textbf{Pre.} & \textbf{Rec.} & \textbf{Acc.} & \textbf{Pre.} & \textbf{Rec.} & \textbf{Acc.} & \textbf{Pre.} & \textbf{Rec.} & \textbf{Acc.} & \textbf{Pre.} & \textbf{Rec.} & \textbf{Acc.} & \textbf{Pre.} & \textbf{Rec.} \\
			\midrule
			\textbf{SID} & 0.347 & 0.580 & 0.347 & 0.324 & 0.570 & 0.324 & 0.382 & 0.313 & 0.382 & 0.373 & 0.678 & 0.373 & 0.290 & 0.604 & 0.290 & 0.264 & 0.502 & 0.264 \\
			\textbf{IVAD} & 0.864 & 0.881 & 0.864 & 0.886 & 0.892 & 0.886 & 0.906 & 0.912 & 0.906 & 0.944 & 0.949 & 0.944 & 0.825 & 0.840 & 0.825 & 0.893 & 0.901 & 0.893 \\
			\bottomrule
		\end{tabular}
	\end{adjustbox}
\end{table*}

\subsection{Experiment 3: Color Augmentation via Bayesian Optimization in the Singular Illumination Dataset}

\begin{figure}[!t]
	\centering
	\includegraphics[width=4.5in]{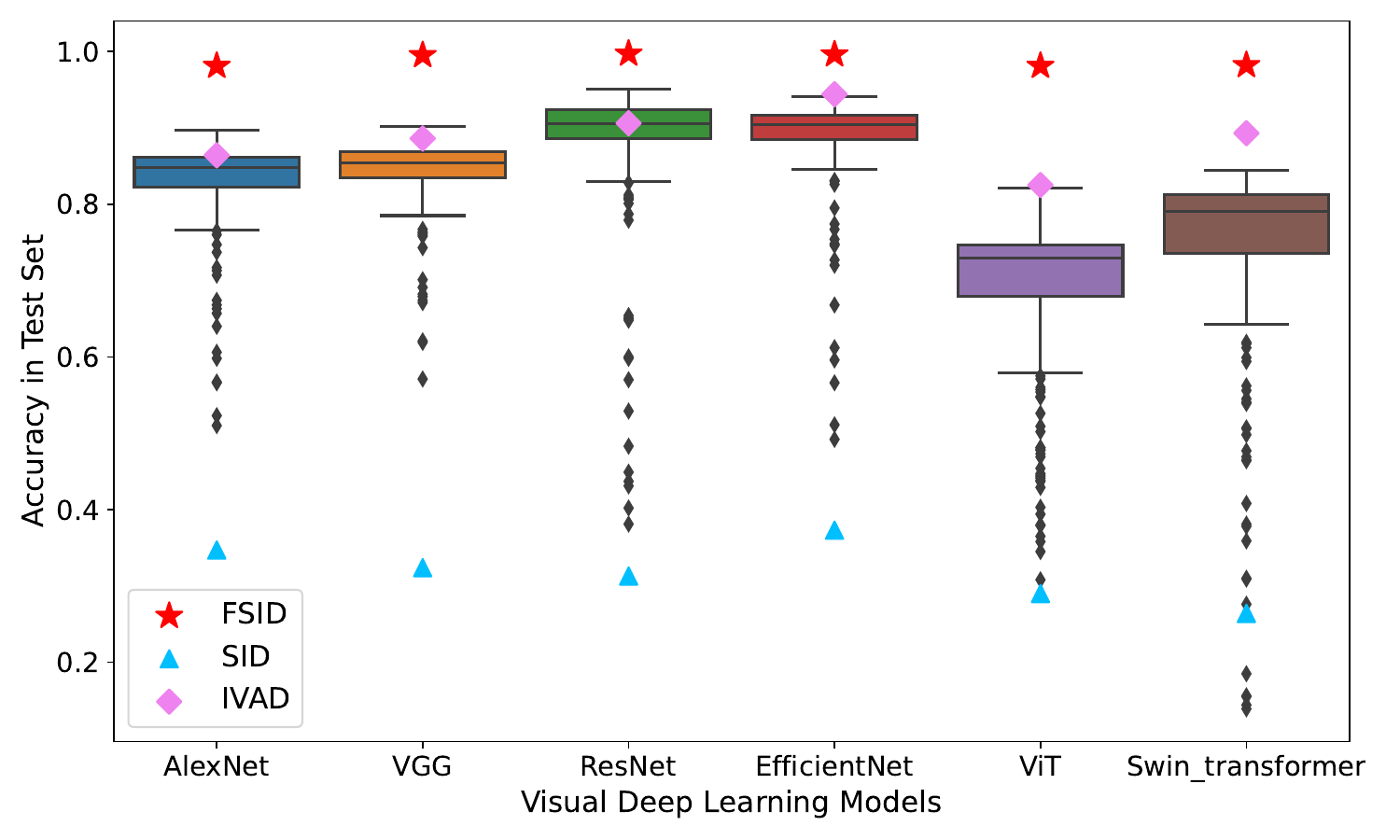}
	\caption{Each box displayed the optimization of color enhancement parameters achieved through Bayesian optimization using Optuna across six distinct visual models over 200 iterations. It highlighted the progress in model generalization due to color-based data augmentation, showing improvements beyond the IVAD's results from Experiment 2, while still emphasizing the gap in performance compared to the FSID dataset's real-world illumination variations.  }
	\label{fig6}
\end{figure}

During the training process of deep learning-based visual models, data augmentation techniques play an important role. In Experiment 2 we demonstrated that color channel enhancement could significantly improve model performance. Given that our study mainly focused on the "illumination" attribute, this experiment also focused on color-based data augmentation methods. We utilized Torchvision, a package from the PyTorch~\cite{ref43} project that provides tools and datasets for computer vision. It provided a color augmentation function "torchvision.transforms.ColorJitter" which included 4 variables: brightness, contrast, saturation, and hue. In this section, we aim to maximize the model's performance on the test set by searching for the best parameter configurations on torchvision.transforms.ColorJitter data augmentation function. 

To precisely adjust these four color enhancement parameters, we employed the Optuna Bayesian optimization framework~\cite{ref44}, with the Tree-structured Parzen Estimator (TPE) as the parameter sampling strategy. During the optimization process, each TPE iteration generated a new set of parameter configurations. These configurations were applied to data augmentation and subsequently used to train models according to the setup established in Experiment 1. After training, we evaluated the effectiveness of each iteration by measuring the model's accuracy on the test set. For six different deep learning models, we conducted 200 optimization iterations for searching the best color augmentation parameter configurations. As shown in Fig.5, the results demonstrated that after 200 iterations of Bayesian optimization for data augmentation, with the best parameter configurations, the models' performance approached or even surpassed the illumination mapping data augmentation methods used in IVAD. Our experiments showed that color enhancement techniques in Torchvision, fine-tuned through Bayesian optimization, could surpass the illumination vector mapping augmentation method in Experiment 2. However, despite all these improvements, a generalization gap still existed when compared to the FSID dataset, which included actual illumination changes. This emphasized the irreplaceable of visual feature complexity in real datasets for constructing a robust model generalization.

\section{Conclusion}

\begin{table*}[ht]
	\centering
	\caption{Performance analysis of models across different distributions and augmentation methods within FSID, SID, IVAD, and Bayesian Optimization Data Augmentation (BO-DA). It demonstrates that while data augmentation methods (IVAD and BO-DA) significantly improved models' generalization, a notable gap persists when compared to FSID which took data from a real-world illumination distribution. The gap highlights that artificial illumination variations introduced through data augmentation inherently involve certain generalization limitations.}
	\label{tab4}
	\begin{adjustbox}{width=1\textwidth,center}
		\begin{tabular}{@{}ccccccccccccccccccc@{}}
			\toprule
			& \multicolumn{3}{c}{\textbf{AlexNet}} & \multicolumn{3}{c}{\textbf{VGG}} & \multicolumn{3}{c}{\textbf{ResNet}} & \multicolumn{3}{c}{\textbf{EfficientNet}} & \multicolumn{3}{c}{\textbf{ViT}} & \multicolumn{3}{c}{\textbf{Swin\_T}} \\
			\cmidrule(lr){2-4} \cmidrule(lr){5-7} \cmidrule(lr){8-10} \cmidrule(lr){11-13} \cmidrule(lr){14-16} \cmidrule(lr){17-19}
			\textbf{Metrics} & \textbf{Acc.} & \textbf{Pre.} & \textbf{Rec.} & \textbf{Acc.} & \textbf{Pre.} & \textbf{Rec.} & \textbf{Acc.} & \textbf{Pre.} & \textbf{Rec.} & \textbf{Acc.} & \textbf{Pre.} & \textbf{Rec.} & \textbf{Acc.} & \textbf{Pre.} & \textbf{Rec.} & \textbf{Acc.} & \textbf{Pre.} & \textbf{Rec.} \\
			\midrule
			FSID & \textbf{0.981} & \textbf{0.982} & \textbf{0.981} & \textbf{0.995} & \textbf{0.995} & \textbf{0.995} & \textbf{0.997} & \textbf{0.997} & \textbf{0.997} & \textbf{0.996} & \textbf{0.996} & \textbf{0.996} & \textbf{0.981} & \textbf{0.982} & \textbf{0.981} & \textbf{0.982} & \textbf{0.983} & \textbf{0.982} \\
			SID & 0.347 & 0.580 & 0.347 & 0.324 & 0.570 & 0.324 & 0.382 & 0.313 & 0.382 & 0.373 & 0.678 & 0.373 & 0.290 & 0.604 & 0.290 & 0.264 & 0.502 & 0.264 \\
			IVAD & 0.864 & 0.881 & 0.864 & 0.886 & 0.892 & 0.886 & 0.906 & 0.912 & 0.906 & 0.944 & 0.949 & 0.944 & 0.825 & 0.840 & 0.825 & 0.893 & 0.901 & 0.893 \\
			BO-DA & 0.897 & 0.900 & 0.897 & 0.902 & 0.908 & 0.902 & 0.951 & 0.952 & 0.951 & 0.941 & 0.943 & 0.941 & 0.821 & 0.822 & 0.821 & 0.844 & 0.855 & 0.844 \\
			\bottomrule
		\end{tabular} 
	\end{adjustbox}
\end{table*}

To conclude our study, we summarized all results in Table 3. In our research, we focused on the visual representation variable "illumination" by forming the Full Spectrum Illumination Dataset (FSID) with uniform distribution and the Singular Illumination Dataset (SID) with singular distribution on datasets of a classification task. In Experiment 1, we proved that when a visual representation variable degenerated to a singular distribution, it will occur a catastrophic decline in deep learning-based visual models. We performed an illumination vector mapping data augmentation method in Experiment 2. Models' generalization abilities trained by the Illumination Vector Augmentation Dataset (IVAD) had significant improvements. Experiment 3 further conducted a Bayesian Optimization Data Augmentation (BO-DA) method, which slightly outperformed the models' performance trained with IVAD.

Nevertheless, whether employing intuitive color mapping techniques or implementing color-based data augmentation through Bayesian optimization, these strategies still exhibit a generalization gap when compared to models trained on a complex, real-world illumination dataset. This outcome highlights that while data augmentation can enhance model generalization, it has inherent limitations. It is essential to ensure that different visual representation variables are sufficiently complex and diverse. Most importantly, any visual representation variable should not degrade into a singular distribution. Proper dataset design is critical for achieving robust model generalization, emphasizing the importance of incorporating diverse visual features.

\section{Data Limitation}
One limitation of our study is the image format. Currently, images were saved in PNG format using an Intel RealSense D435i camera, which does not preserve physical illumination information as effectively as RAW format. RAW captures linear data directly from the sensor, crucial for accurate color correction and augmentation. We acknowledge that using PNG may affect our results and plan to use RAW format in future work to enhance data accuracy and robustness.

%
%
%
%

\end{document}